\documentclass[journal]{IEEEtran}
\ifCLASSINFOpdf
   \usepackage[pdftex]{graphicx}
\else
\fi
\usepackage{microtype}

\usepackage{amsmath}
\usepackage{physics}
\usepackage{amsmath}
\usepackage{tikz}
\usepackage{mathdots}
\usepackage{yhmath}
\usepackage{cancel}
\usepackage{color}
\usepackage{siunitx}
\usepackage{array}
\usepackage{multirow}
\usepackage{amssymb}
\usepackage{gensymb}
\usepackage{tabularx}
\usepackage{extarrows}
\usepackage{booktabs}
\usetikzlibrary{fadings}
\usetikzlibrary{patterns}
\usetikzlibrary{shadows.blur}
\usetikzlibrary{shapes}
\usepackage{lipsum,adjustbox}

\begin{document}
\title{Synthetic Dataset Generation and Learning From Demonstration Applied to Industrial Manipulation}

\author{Alireza~Barekatain,
        Hamed~Rahimi~Nohooji,~\IEEEmembership{Member,~IEEE,}
        and~Holger~Voos,~\IEEEmembership{Member,~IEEE}% <-this % stops a space
\thanks{Authors are affiliated with University of Luxembourg, Interdisciplinary Center for Security, Reliability, and Trust (SnT), Luxembourg. e-mail: alireza.barekatain@uni.lu}% <-this % stops a space
}

\maketitle

% \begin{abstract}
% The abstract goes here.
% \end{abstract}

% \begin{IEEEkeywords}
% IEEE, IEEEtran, journal, \LaTeX, paper, template.
% \end{IEEEkeywords}

% \fi
\IEEEpeerreviewmaketitle

\section{Introduction}\label{intro}
Shifting from a fixed production scheme to flexible production, a concept known as Mass Customization, raises nontrivial uncertainties when it comes to robotic automation. One source of uncertainty comes from the fact that the robot system will be in an unstructured environment, meaning that the robot’s perception needs to perform a more sophisticated task. As a common example, conveyer belts remove the burden of the vision system by eliminating uncertainties of objects’ pose, while in a flexible manipulation setting, the robot’s perception needs to locate a variety of different industrial objects in a tray. One other source of uncertainty comes from the ever-changing nature of desired tasks. It means that the robot is not fixed to a set of predefined trajectories. As opposed to a fixed scheme, there is no longer an absolute path to follow. Robot needs to devise a new motion relative to the location of the detected objects and the required task to perform on that specific object. The variety of possible tasks and motions introduces a new type of uncertainty into the system. In this work, we are aiming to move towards tackling the challenges of the aforementioned uncertainties.

The uncertainty of vision systems becomes more challenging when the objects are metallic, shiny, and symmetric, which is common in industry. The raised challenges are concerned with the pose estimation problem, where a vision device (e.g. a camera) estimates the position and orientation of an object with respect to a reference frame. The featureless nature of such objects leads us to focus on deep-learning (DL)-based methods, rather than conventional approaches. However, the process of creating a dataset to train a DL model is time consuming and requires a specific setup, which makes the whole process inefficient in terms of required effort. On the other hand, as the robot tasks are not fixed, a robotic expert needs to manually modify the program to match the required flexibility of production. Such bottlenecks in a flexible robotic assembly line would lead to inefficiencies and prove ineffective in the long run.

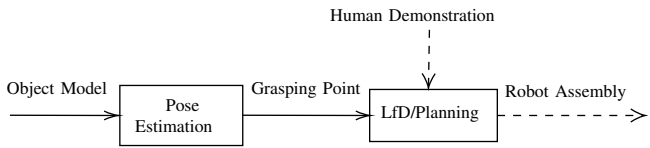
\begin {figure}[h]%[!hbtp]
 \centering
  \begin{adjustbox}{width=\textwidth*19/40}

\tikzset{every picture/.style={line width=0.75pt}} %set default line width to 0.75pt        

\begin{tikzpicture}[x=0.75pt,y=0.75pt,yscale=-1,xscale=1]
 
%Straight Lines [id:da28347712058597496] 
\draw    (8.73,152.32) -- (88.16,152.5) ;
\draw [shift={(90.16,152.5)}, rotate = 180.13] [color={rgb, 255:red, 0; green, 0; blue, 0 }  ][line width=0.75]    (10.93,-3.29) .. controls (6.95,-1.4) and (3.31,-0.3) .. (0,0) .. controls (3.31,0.3) and (6.95,1.4) .. (10.93,3.29)   ;
%Shape: Rectangle [id:dp9311221548800159] 
\draw   (91,130) -- (183,130) -- (183,175) -- (91,175) -- cycle ;
%Straight Lines [id:da47922216302447707] 
\draw    (182.73,152.32) -- (276.55,152.5) ;
\draw [shift={(278.55,152.5)}, rotate = 180.11] [color={rgb, 255:red, 0; green, 0; blue, 0 }  ][line width=0.75]    (10.93,-3.29) .. controls (6.95,-1.4) and (3.31,-0.3) .. (0,0) .. controls (3.31,0.3) and (6.95,1.4) .. (10.93,3.29)   ;
%Shape: Rectangle [id:dp213972983459354] 
\draw   (278,133) -- (373.55,133) -- (373.55,172) -- (278,172) -- cycle ;
%Straight Lines [id:da35736119238888575] 
\draw  [dash pattern={on 4.5pt off 4.5pt}]  (322.73,88.32) -- (322.03,130) ;
\draw [shift={(322,132)}, rotate = 270.95] [color={rgb, 255:red, 0; green, 0; blue, 0 }  ][line width=0.75]    (10.93,-4.9) .. controls (6.95,-2.3) and (3.31,-0.67) .. (0,0) .. controls (3.31,0.67) and (6.95,2.3) .. (10.93,4.9)   ;
%Straight Lines [id:da14706676861609203] 
\draw  [dash pattern={on 4.5pt off 4.5pt}]  (373.36,152.5) -- (482.36,152.5) ;
\draw [shift={(484.36,152.5)}, rotate = 180] [color={rgb, 255:red, 0; green, 0; blue, 0 }  ][line width=0.75]    (10.93,-4.9) .. controls (6.95,-2.3) and (3.31,-0.67) .. (0,0) .. controls (3.31,0.67) and (6.95,2.3) .. (10.93,4.9)   ;

% Text Node
\draw (138,152.5) node   [align=left] {\begin{minipage}[lt]{53.16pt}\setlength\topsep{0pt}
\begin{center}
Pose
\end{center}
 Estimation
\end{minipage}};
% Text Node
\draw (285,144) node [anchor=north west][inner sep=0.75pt]   [align=left] {LfD/Planning};
% Text Node
\draw (5,126) node [anchor=north west][inner sep=0.75pt]   [align=left] {Object Model};
% Text Node
\draw (188,126) node [anchor=north west][inner sep=0.75pt]   [align=left] {Grasping Point};
% Text Node
\draw (247,71) node [anchor=north west][inner sep=0.75pt]   [align=left] {Human Demonstration};
% Text Node
\draw (378,128) node [anchor=north west][inner sep=0.75pt]   [align=left] {Robot Assembly};

\end{tikzpicture}

  \end{adjustbox}
	\caption{The overview of the automated industrial manipulation pipeline.}\label{figoverall}
\end{figure}

The aim of this study is to investigate an automated industrial manipulation pipeline, where assembly tasks can be flexibly adapted to the production without the need of a robotic expert, both for the vision system and the robot program. The objective of this study is first, to develop a synthetic- dataset-generation pipeline \cite{denninger2019blenderproc} with special focus on industrial parts, and second, to use Learning-from-Demonstration (LfD) methods to replace manual robot programming, so that a non-robotic expert/process engineer can introduce a new manipulation task by teaching it to the robot \cite{ravichandar2020recent}. The overview of the explained pipeline is shown in Fig. \ref{figoverall}.

\begin{verbatim}
dataset-output:
    - pose: pose0.npy, pose1.npy, ...
    - rgb: 0.jpg, 1.jpg, ...
    - mask: 0.png, 1.png, ...

\end{verbatim}
\vspace{-5mm}

\begin{figure}[t]
\begin{center}
\includegraphics[width = 0.29 \textwidth]{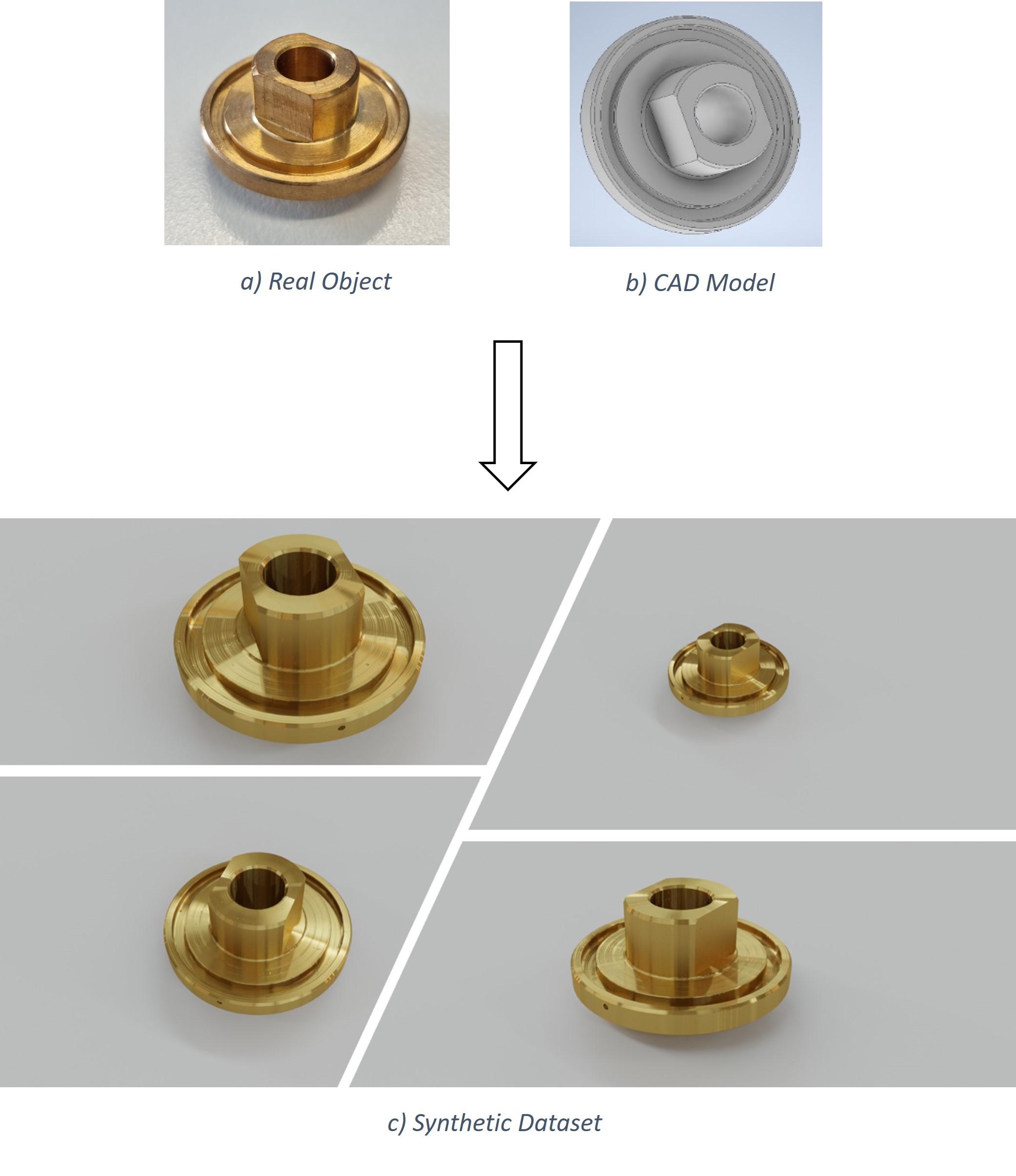}
\caption{An example of synthetic dataset generation. (a) and (b) show the input to the process, where the CAD file is used to recreate the geometry and shaders are customized based on the real appearance and material of the object, resulting in (c). }
\label{figobject}
\end{center}
\vspace{-5mm}
\end{figure}

\section{Pose Estimation}\label{pse}
In this section, our main assumption is that the CAD model of the desired object is available, as it is the case for most industrial cases. The idea is to create a synthetic scene of the object, which is rendered from different viewing angles and is associated with the respective pose of the object in each scene. An example of this procedure is illustrated in Fig. \ref{figobject}. The desired dataset output comprises of object's pose, scene's image, and object's mask in the scene, with the following template:

The main input to our synthetic dataset generation process is the CAD file, which is imported in Blender, a computer graphics software, to create the basic scene.  To enhance the data augmentation of the dataset, the scene can be modified by adding different backgrounds and various objects. Once the basic scene is ready, it is required to add special shading techniques to represent the reflective and metallic appearance of the real-world object. We have developed and applied a shader that replicates the real appearances of metallic parts. The photorealistic scene is created by adding the aforementioned shader and adjusting the color based on the object’s material.

In the next step, a virtual camera renders images from different angles, and  calculates the pose of the target object with respect to the camera frame. We have utilized BlenderProc \cite{denninger2019blenderproc}, a procedural Blender pipeline for photorealistic rendering, to automate this process. To tackle the issue of symmetry, the camera angles are adjusted based on the symmetry lines of the object to avoid confusion during training of the DL model. The final output is a photorealistic dataset which can be used for training any pose estimation method.

\begin{figure}[t]

\begin{center}
\includegraphics[width = 0.25 \textwidth]{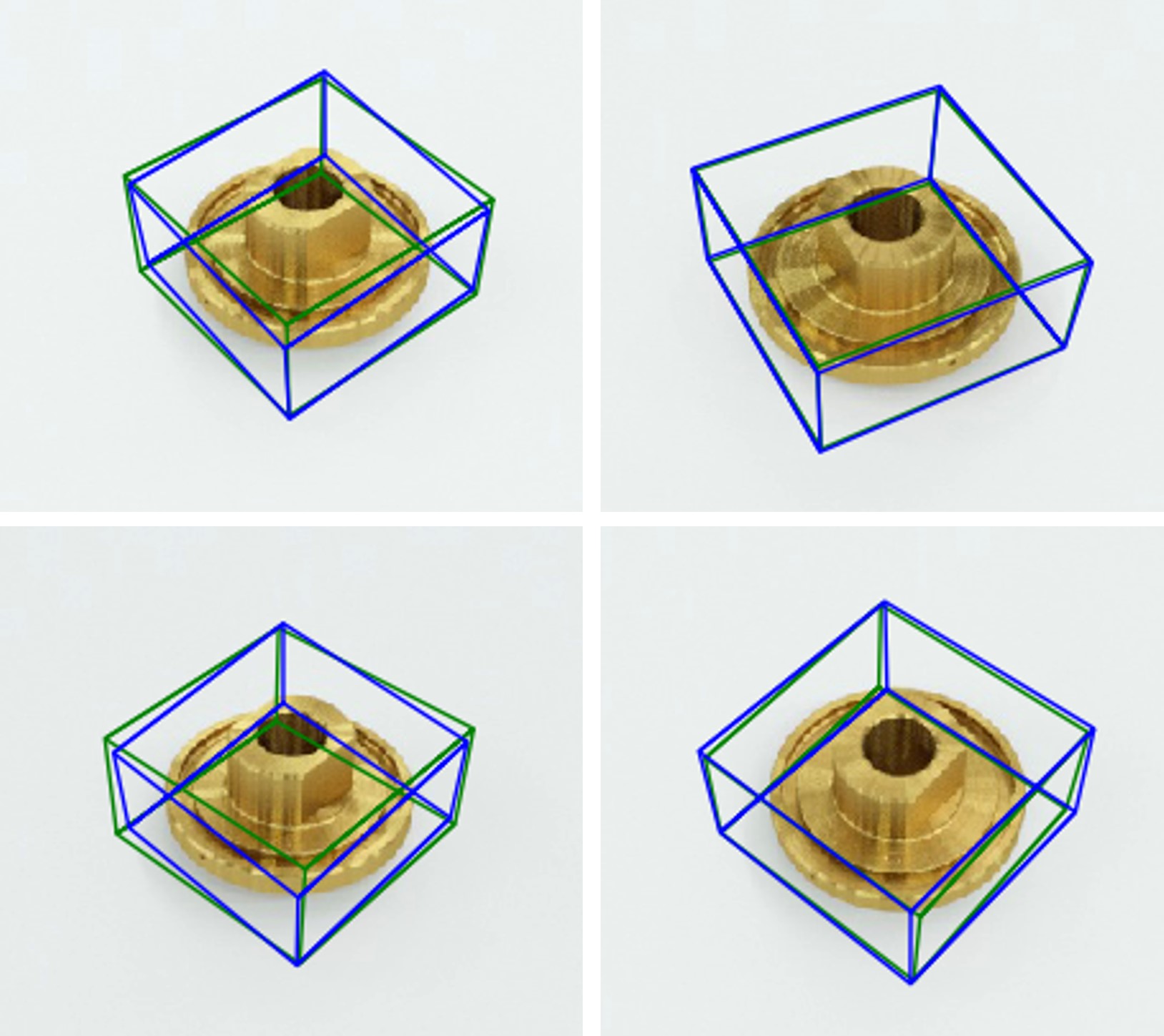}
\caption{Samples from the validation set of a trained PVNet. The green bounding box is the ground-truth pose and the blue bounding box is the estimated pose.}
\label{figpvnet}
\end{center}
\vspace{-5mm}
\end{figure}

As a case study, we have considered a round symmetric industrial object made of brass and with 22mm diameter (Fig. \ref{figobject}. The background scene is chosen to be plain white, and the shader is customized based on the target object’s color. The output dataset has 500 images, which is subsequently used to train a state-of-the-art pose estimation method, namely PVNet \cite{peng2019pvnet}. A number of example outputs of the validation set is shown in Fig. \ref{figpvnet}. By acquiring the pose, it is straightforward to extract the grasping points of the object and feed it to the robot manipulator to perform the task.

\section{Learning from Demonstration}
In this section, we explain how the concept of LfD is applied to teach the robot how to perform a task, instead of manually programming it. The robot has to learn the concept of motion, rather than fixed trajectory points. With the pose estimation methodology explained in Section \ref{pse}, we assume that the robot knows where the grasping points are. Here, the focus is to teach the robot how to perform a task made of a sequence of motions. It is worth mentioning that the starting point of the task depends on the coordinates and the grasping points of the target object to be manipulated, hence there are no fixed trajectory points.

For the LfD approach, we focus on Kinesthetic Teaching \cite{ravichandar2020recent}, where the robot is in a compliant mode and the human operator teaches the robot by moving it around. The core of our LfD is a Dynamic Movement Primitive (DMP), shown in Eqs. \eqref{eq1} and \eqref{eq2}, where the force $f(x)$ is learned to reach goal $g$ using Locally Weighted Regression (LWR) method \cite{ijspeert2013dynamical}.

\begin{equation} \label{eq1}
\tau^2 \ddot{y}=\alpha_z\left(\beta_z(g-y)-\tau \dot{y}\right)+f(x)
\end{equation}
\begin{equation} \label{eq2}
    \tau \dot{x}=-\alpha_x x
\end{equation}

We have developed a Robot-Operating-System (ROS)-based LfD interface where the human operator can teach an indefinite sequence of motions (sub-task) to define a task, while the joint states of the robot is recorded and stored. At the time of operation, the motion sequence is retrieved to train the DMP, and plan a new task motion based on the starting configuration received from the vision system.

\begin{figure}[h]
\begin{center}
\includegraphics[width = 0.29 \textwidth]{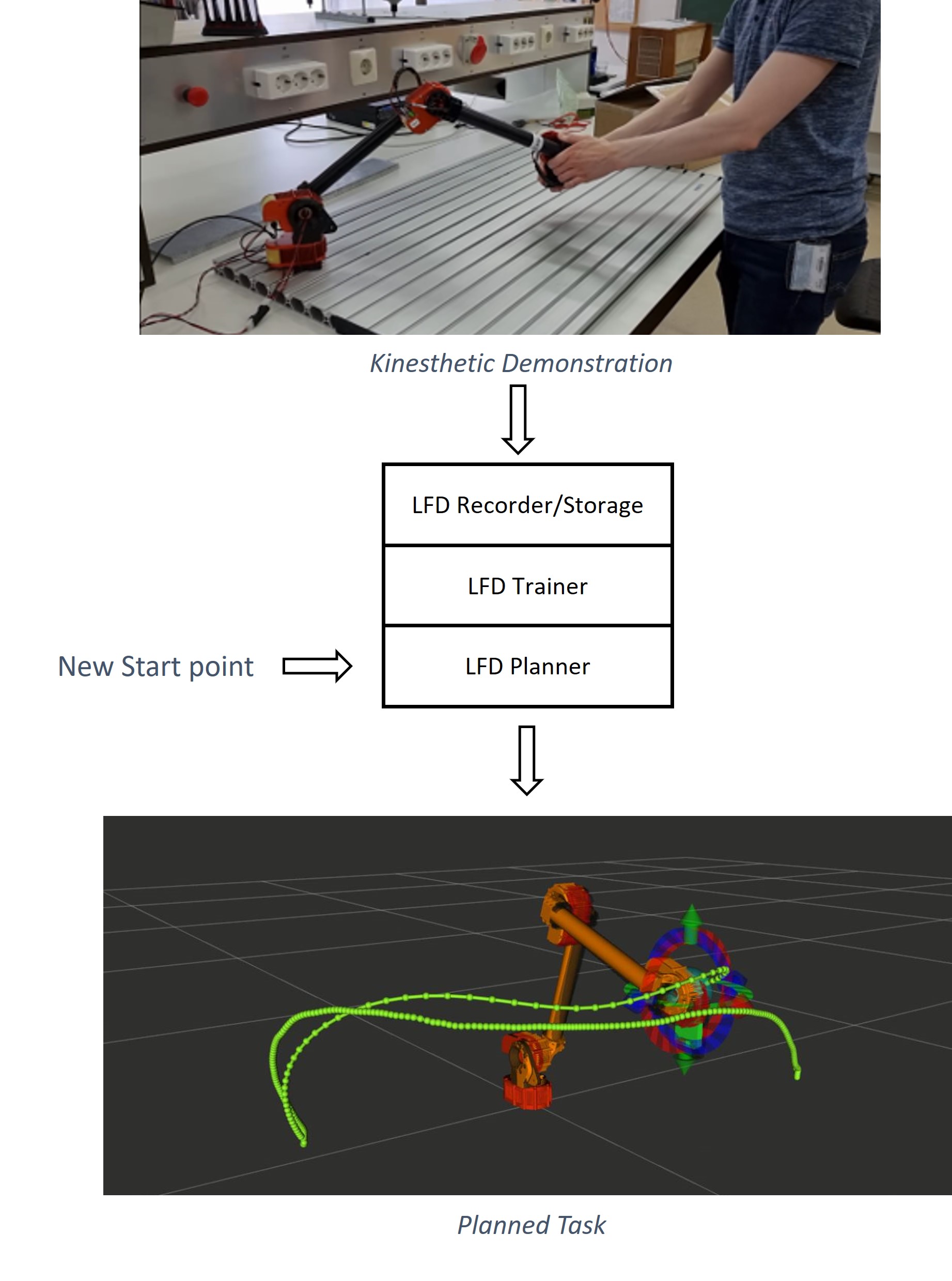}
\caption{The overall workflow of teaching a pick-and-place task. The human trains the method by guiding the robot through the desired movement (dotted trajectory). LfD planner creates a new trajectory from the received starting point to the desired end point (dash-dotted trajectory)}
\label{figlfd}
\end{center}
\vspace{-5mm}
\end{figure}

Here we have considered a pick-and-place motion as a case study. The human operator moves the robot from an arbitrary start point and guides it through the desired motion. After training, the robot has learned to replicate the same motion pattern from a new starting configuration, i.e. the new location of the part. The overall workflow is depicted in Fig. \ref{figlfd}.

% or


\section*{Acknowledgment}

This work is supported by FNR "Fonds national de
la Recherche" (Luxembourg) through Industrial Fellowship Ph.D. grant (ref. 15882013).

\ifCLASSOPTIONcaptionsoff
  \newpage
\fi



%\bibliographystyle{IEEEtran}


% \end{thebibliography}
\bibliographystyle{IEEEtran}
\bibliography{refs.bib}

\begin{thebibliography}{1}
\providecommand{\url}[1]{#1}
\csname url@samestyle\endcsname
\providecommand{\newblock}{\relax}
\providecommand{\bibinfo}[2]{#2}
\providecommand{\BIBentrySTDinterwordspacing}{\spaceskip=0pt\relax}
\providecommand{\BIBentryALTinterwordstretchfactor}{4}
\providecommand{\BIBentryALTinterwordspacing}{\spaceskip=\fontdimen2\font plus
\BIBentryALTinterwordstretchfactor\fontdimen3\font minus
  \fontdimen4\font\relax}
\providecommand{\BIBforeignlanguage}[2]{{%
\expandafter\ifx\csname l@#1\endcsname\relax
\typeout{** WARNING: IEEEtran.bst: No hyphenation pattern has been}%
\typeout{** loaded for the language `#1'. Using the pattern for}%
\typeout{** the default language instead.}%
\else
\language=\csname l@#1\endcsname
\fi
#2}}
\providecommand{\BIBdecl}{\relax}
\BIBdecl

\bibitem{denninger2019blenderproc}
M.~Denninger, M.~Sundermeyer, D.~Winkelbauer, Y.~Zidan, D.~Olefir,
  M.~Elbadrawy, A.~Lodhi, and H.~Katam, ``Blenderproc,'' \emph{arXiv preprint
  arXiv:1911.01911}, 2019.

\bibitem{ravichandar2020recent}
H.~Ravichandar, A.~S. Polydoros, S.~Chernova, and A.~Billard, ``Recent advances
  in robot learning from demonstration,'' \emph{Annual review of control,
  robotics, and autonomous systems}, vol.~3, pp. 297--330, 2020.

\bibitem{peng2019pvnet}
S.~Peng, Y.~Liu, Q.~Huang, X.~Zhou, and H.~Bao, ``Pvnet: Pixel-wise voting
  network for 6dof pose estimation,'' in \emph{Proceedings of the IEEE/CVF
  Conference on Computer Vision and Pattern Recognition}, 2019, pp. 4561--4570.

\bibitem{ijspeert2013dynamical}
A.~J. Ijspeert, J.~Nakanishi, H.~Hoffmann, P.~Pastor, and S.~Schaal,
  ``Dynamical movement primitives: learning attractor models for motor
  behaviors,'' \emph{Neural computation}, vol.~25, no.~2, pp. 328--373, 2013.

\end{thebibliography}


% \end{IEEEbiography}


%\newpage



%\vfill


\end{document}